\documentclass[10pt,twocolumn,letterpaper]{article}

\usepackage{iccv}              

\usepackage{multirow}
\usepackage{colortbl}

\definecolor{iccvblue}{rgb}{0.21,0.49,0.74}
\usepackage[pagebackref,breaklinks,colorlinks,allcolors=iccvblue]{hyperref}

\def\paperID{11004} 
\def\confName{ICCV}
\def\confYear{2025}

\title{One-Shot Knowledge Transfer for Scalable Person Re-Identification}

\author{
Longhua Li$^{1,2}$ \quad Lei Qi$^{1,2}$\thanks{Corresponding authors.} \quad Xin Geng$^{1,2}$\footnotemark[1] \\
$^1$School of Computer Science and Engineering, Southeast University, Nanjing, China\\
$^2$Key Laboratory of New Generation Artificial Intelligence Technology and Its Interdisciplinary\\Applications (Southeast University), Ministry of Education, China\\
{\tt\small \{lhli, qilei, xgeng\}@seu.edu.cn}
}

\begin{document}
\maketitle
\begin{abstract}
Edge computing in person re-identification (ReID) is crucial for reducing the load on central cloud servers and ensuring user privacy. Conventional compression methods for obtaining compact models require computations for each individual student model. When multiple models of varying sizes are needed to accommodate different resource conditions, this leads to repetitive and cumbersome computations.
To address this challenge, we propose a novel knowledge inheritance approach named OSKT (One-Shot Knowledge Transfer), which consolidates the knowledge of the teacher model into an intermediate carrier called a weight chain. When a downstream scenario demands a model that meets specific resource constraints, this weight chain can be expanded to the target model size without additional computation.
OSKT significantly outperforms state-of-the-art compression methods, with the added advantage of one-time knowledge transfer that eliminates the need for frequent computations for each target model.
\end{abstract}
    
\section{Introduction}
\label{sec:intro}

\begin{figure}[t]
    \centering
    \begin{subfigure}{\linewidth}
        \centering
        \includegraphics[width=\linewidth]{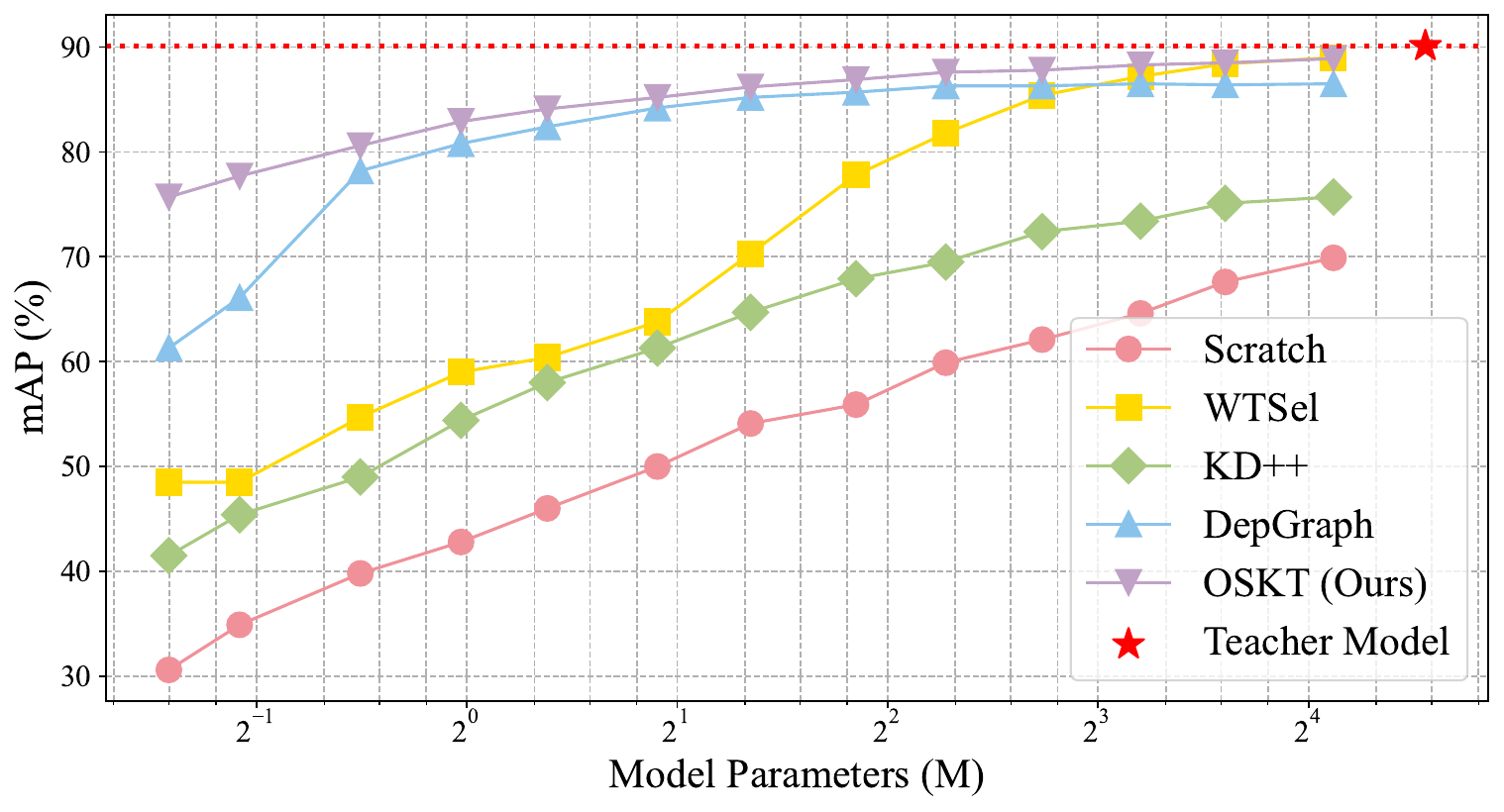}
        \caption{mAP vs. Model Size (Log Scale) on Market1501}
        \label{fig:score_param}
    \end{subfigure}

    \vspace{2mm} 

    \begin{subfigure}{\linewidth}
        \centering
        \includegraphics[width=\linewidth]{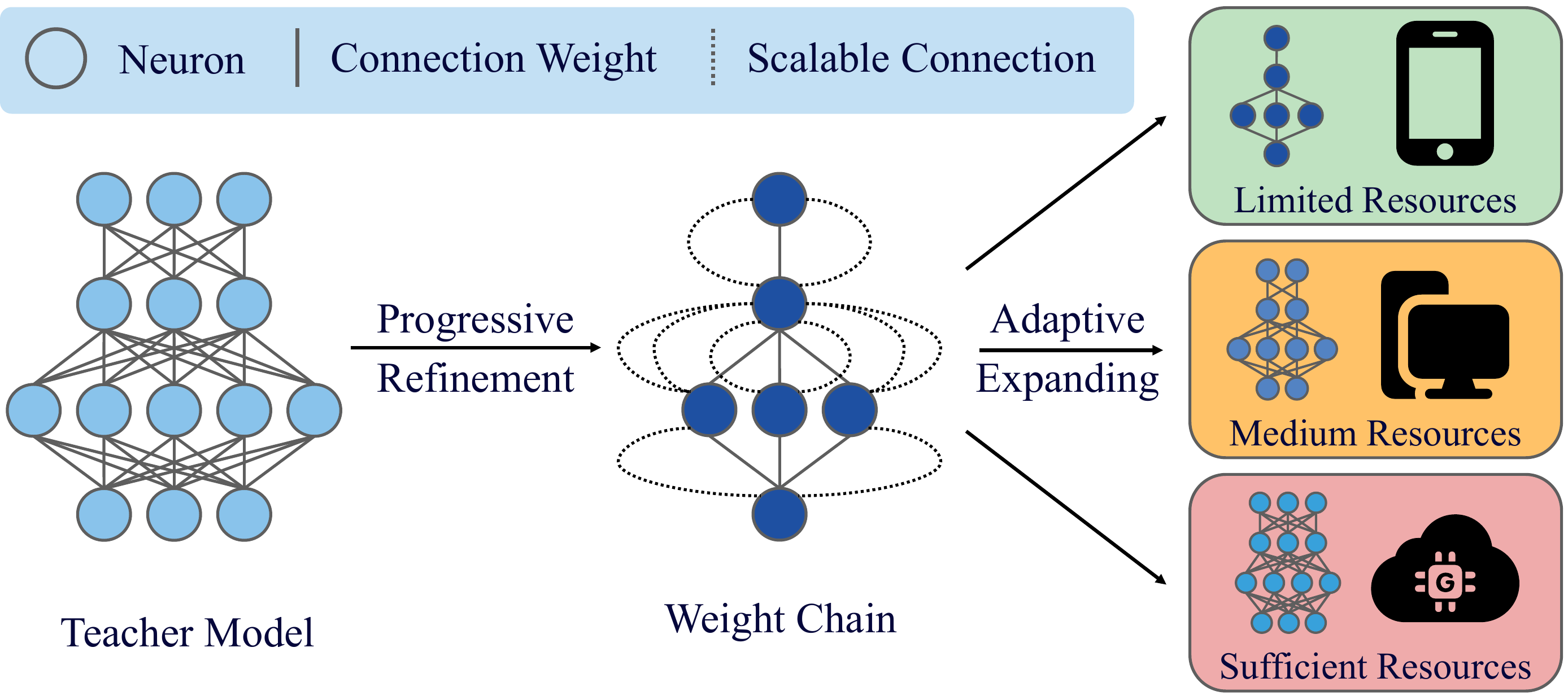}
        \caption{Conceptual Diagram of the Proposed Method}
        \label{fig:overview}
    \end{subfigure}
    \caption{(a) Transfer knowledge from a pre-trained ResNet50 to student models of different sizes. (b) Refine knowledge into a weight chain in a single pass, enabling adaptive expansion of models of varying sizes meeting downstream resource constraints.}
    \label{fig_cluster_result}
\end{figure}




Person re-identification (ReID) is pivotal for surveillance and smart city systems \cite{behera2020person}, aiming to uniquely identify individuals across disparate camera views \cite{zheng2016person,ye2021deep}. While deploying ReID models on edge devices is crucial for mitigating cloud computing burdens and safeguarding data privacy \cite{zhang2022learnable,ye2024securereid}, recent advancements in the ReID field are often encumbered by heavy parameter loads and intensive computational requirements \cite{fu2022large,li2023clip,dou2023identity}, restricting their applicability in resource-constrained scenarios.

To achieve an optimal accuracy-cost trade-off, different edge computing scenarios require carefully customized models to maximize re-identification performance within their resource constraints. A common approach for obtaining a high-performance compact model is to compress a large, well-trained teacher model into a smaller student model through techniques such as pruning \cite{lecun1989optimal} and distillation \cite{hinton2015distilling}. Nevertheless, these methods are typically limited to generating a single model of a fixed size at a time, requiring separate and often intricate processes for each downstream scenario, thus consuming significant resources.

In this work, we propose a knowledge inheritance method suitable for general model architectures. It requires only a single computation to obtain student models of various sizes, without the need for the repetitive and computationally expensive training procedures typically required by traditional compression methods for each target size. The knowledge of the teacher model is first refined into a weight chain, which acts as an intermediate knowledge carrier. When a student model of a specific size is required, the weight chain can be rapidly expanded to the target size without additional computation. The resulting model can be directly deployed or fine-tuned for downstream tasks.

Specifically, the weight chain maintains the same depth (number of layers) as the teacher model, with each layer's weights being refined from the corresponding layer in the teacher model. However, its width (feature dimension) is substantially reduced compared to the teacher model, while maintaining the flexibility to be expanded to generate models of any intermediate width between the weight chain and the teacher model. Notably, the weight chain preserves the complete normalization layers of the original width to effectively reuse the core feature dimensions of its output, thereby providing richer feature representations. When a model of specific size is required, this can be achieved by stacking the weight matrix rows of the weight chain and correspondingly merging the affine transformation parameters of the normalization layers to attain the desired width.

The process of refining the weight chain from the teacher model involves two key steps: initialization and progressive refinement of the weight chain.
Initially, we set the weight matrix rows in the weight chain as cluster centers from the teacher model's corresponding rows, as these centers exhibit functional representativeness and facilitate subsequent refinement processes.
For progressive refinement, we jointly train the teacher model and the smallest student model (with width matching the weight chain) built by the weight chain, propagating gradients back to the teacher model and the weight chain. Serving as a bridge between the teacher model and all student models, this approach enables simultaneous training of both ends while implicitly training intermediate models, eliminating the need for intensive training of individual student models of varying sizes. Furthermore, we incorporate an MSE loss between the weight chain rows and the corresponding row clusters of the teacher model to ensure that all weight rows within the same cluster are refined into a single core weight row.
Ultimately, the refined weight chain can be expanded to generate models of any size between its own width and that of the teacher model without additional computation.

Our contributions are summarized as follows:

\begin{itemize}
    \item To our knowledge, we are the first to propose a one-shot computation-based method for generating scalable person ReID models, adaptable to varying resource scenarios.
    \item Our method can be seamlessly integrated with general model architectures, providing highly accurate models that outperform traditional knowledge transfer methods.
    \item Extensive experiments demonstrate that our approach consistently performs well across diverse real-world scenarios, providing a robust solution for person ReID tasks.
\end{itemize}

\section{Related Work}
\label{sec:related_work}

\subsection{Person Re-Identification}
In Person ReID, early methods \cite{zheng_scalable_2015,liao2015person,zheng2016person} focused on handcrafted features and metric learning techniques \cite{yi2014deep}. With the rise of deep learning, Convolutional Neural Network (CNN)-based approaches \cite{cheng2016person,sun2018beyond,wang2018learning,luo2019bag,xu2024distribution} dominated for a long time, significantly boosting ReID performance. Recently, Vision Transformer (ViT) 
 \cite{dosovitskiy2020image}-based methods \cite{he2021transreid,luo2021self,zhu2022pass,he2024instruct,zhang2024view,guo2024lidar,qin2024noisy} have shown promising results in capturing cross-view relationships.
However, these methods are developed on mainstream base models and are not suitable for deployment in resource-constrained edge nodes in practical applications. Recent works \cite{zhouOmniScaleFeatureLearning2019,liCombinedDepthSpace2021,guMSINetTwinsContrastive2023} have designed lightweight models for person ReID, but these still require extensive pre-training on large-scale datasets like ImageNet \cite{deng2009imagenet} and LUPerson \cite{fu2021unsupervised} to achieve high performance. Therefore, achieving high performance in different resource-limited scenarios still involves repetitive computation. In contrast, our method refines the knowledge from the teacher model into a weight chain through a one-time computation. When a model of a specific size is required, the weight chain can be instantly expanded to the target-sized model without training.

\subsection{Knowledge Transfer}
Knowledge transfer involves techniques that allow transferring knowledge from one model to another, enabling efficient model initialization using teacher models. We categorize these methods into two types: those that use additional guidance from teacher models and those based solely on the weights of teacher models. The first type is primarily represented by knowledge distillation \cite{hinton2015distilling,wang2023improving}, while the second includes commonly used methods such as model pruning \cite{lecun1989optimal,wen2016learning,fang_depgraph_2023}. These methods often require costly distillation or pruning with large pre-training datasets, leading to significant computational overhead when initializing scenario-specific models.  
Other methods in the second category include Net2Net \cite{chenNet2NetAcceleratingLearning2016}, DPIAT \cite{czyzewskiBreakingArchitectureBarrier2022}, Weight Selection \cite{xuInitializingModelsLarger2023}, Weight Distillation \cite{linWeightDistillationTransferring2021}, Learngene \cite{wang2022learngene} and so on. However, these methods are either not flexible enough to generate densely-sized student models or inefficient at transferring the knowledge of teacher models, requiring extensive computations to produce student models of various sizes.
Our method ingeniously leverages the weights of the teacher model and incorporates a sophisticated one-shot training strategy to derive densely-sized student models.

\section{Methodology}

To efficiently generate size-specific models for person ReID across various scenarios, we propose a novel framework termed OSKT. As shown in Figure \ref{fig:method_process}, OSKT consists of two core components: (1) weight chain refinement from the teacher model and (2) student model generation via the weight chain. This section is organized as follows: First, we unify CNN and ViT architectures for clarity. Then we formally introduce the weight chain concept, followed by a comprehensive technical elaboration of both components.


\subsection{Unified Representation of CNNs and ViTs}
Our framework is universally applicable to mainstream network architectures, including both CNNs and ViTs. To establish architectural unification, we focus on the feature \textit{dimension} (termed as \textit{channel} in CNNs), which serves as a common bridge between these architectures. Figure \ref{fig:cnn_vit_uniform} provides a conceptual visualization of this unified representation.
In CNNs, convolutional filters each generate a distinct output feature channel, while in ViTs, each weight matrix row produces a specific output feature dimension. We abstract these feature-generating units as \textit{rows}. When features are propagated to the next layer, each dimension is independently processed by a corresponding CNN channel or ViT weight matrix column. We abstract these feature-processing units as \textit{columns}.
Both architectures employ normalization layers (e.g., BN in CNNs, LN in ViTs) between layers, where each feature dimension is associated with affine transformation parameters $(\gamma, \beta)$ that scale and shift normalized features, enhancing their representational capacity.

Consider a teacher model with $L$ layers, where $N_{l-1}$ and $N_{l}$ represent the input and output feature dimensions of the $l$-th layer, respectively.
Let $\mathcal{F}_{l, j}$ represent the $j$-th row of the $l$-th layer, where $\mathcal{F}_{l, j} \in \mathbb{R}^{N_{l-1} \times O_{l}}$, with $O_{l}$ denoting the number of operation parameters in the $l$-th layer. For example, in convolutional layers, $O_{l} = K \times K$ (where $K$ is the kernel size), whereas in fully connected layers, $O_{l} = 1$. Then the weight matrices in the teacher model can be parameterized as $\left\{\mathcal{F}_{l} \in \mathbb{R}^{N_{l} \times N_{l-1} \times O_{l}}, 1 \leq l \leq L \right\}$.

\subsection{Weight Chain as the Knowledge Carrier}
Building on the observation of feature dimension redundancy \cite{ayinde2019redundant,chavan2022vision,li2023scconv}, we propose using refined rows from the teacher model's weight matrices as knowledge carriers. The weight chain preserves the teacher model's depth while containing a significantly reduced number of refined rows per layer compared to the teacher's width.
Formally, the weight chain is parameterized as $\left\{\mathcal{F}^{C}_{l} \in \mathbb{R}^{M_{l} \times N_{l-1} \times O_{l}}, 1 \leq l \leq L \right\}$, where $M_{l}$ denotes the number of refined rows in the $l$-th layer. The normalization layers in the weight chain share weights with their teacher counterparts, enabling multiple $(\gamma, \beta)$ pairs to reuse the core feature dimensions generated by the corresponding refined rows. 
By leveraging the weight chain as an intermediate knowledge carrier, our method is built on three key insights:

\begin{figure}[t]
    \centering

    \begin{subfigure}{\linewidth}
        \centering
        \includegraphics[width=\linewidth]{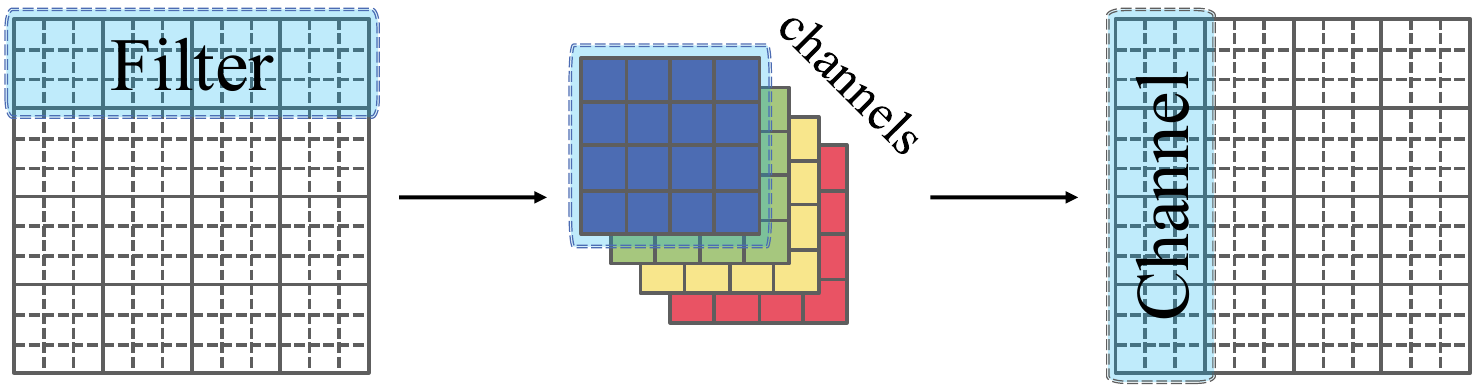}
        \caption{Filter-Channel Connection via Feature Map}
        \label{fig:cnn_rep}
    \end{subfigure}
    
    \vspace{2mm} 

    \begin{subfigure}{\linewidth}
        \centering
        \includegraphics[width=\linewidth]{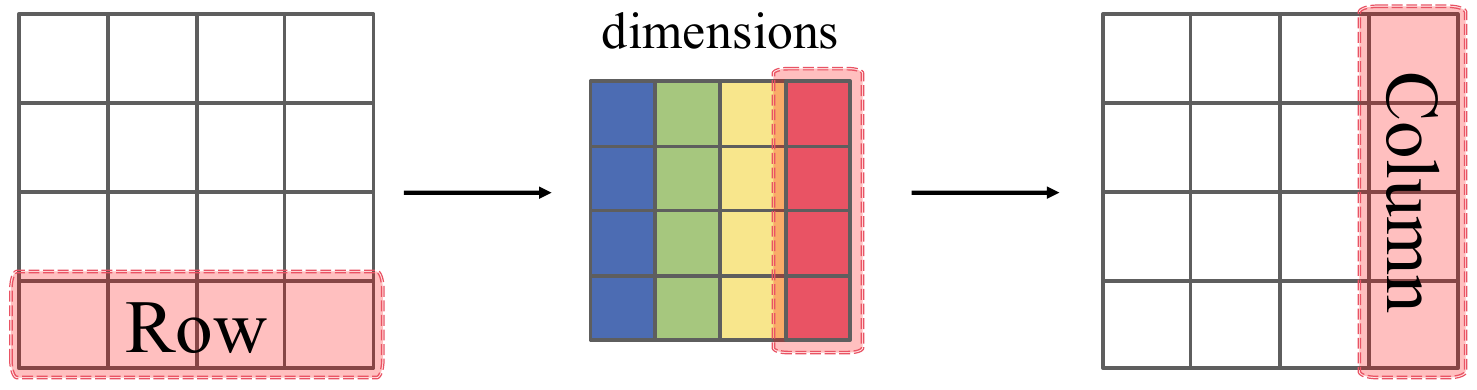}
        \caption{Row-Column Connection via Feature Dimension}
        \label{fig:vit_rep}
    \end{subfigure}

    \caption{(a) In CNNs, a filter is abstracted as a \textit{row}, while a channel in the weight matrix is abstracted as a \textit{column}. (b) In ViTs, the weight matrix is also abstracted into \textit{row}s and \textit{column}s.}
    \label{fig:cnn_vit_uniform}
\end{figure}

\begin{figure}[t]
    \centering
    \includegraphics[width=\linewidth]{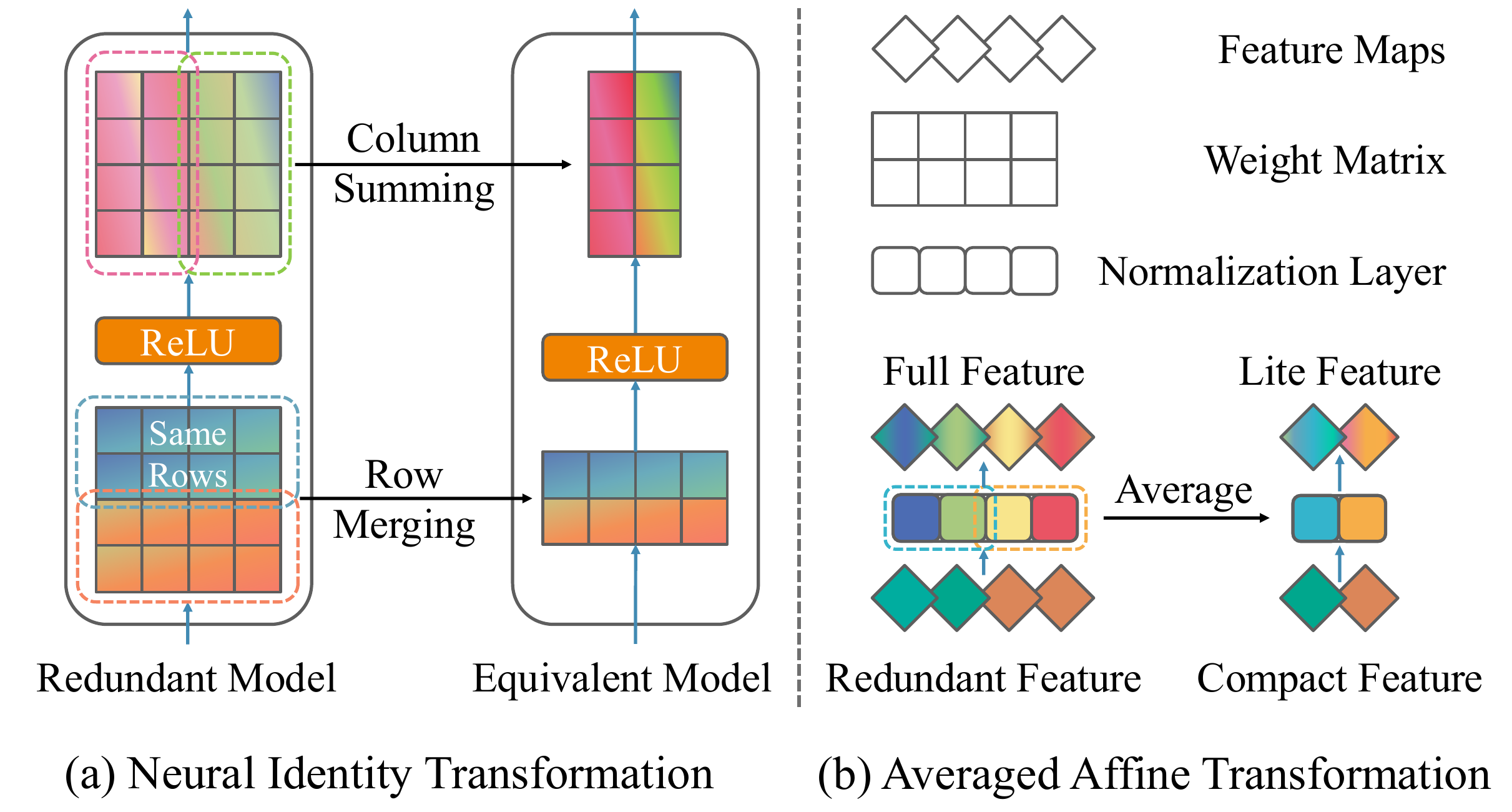}
    \caption{(a) Achieving identity transformation by merging identical rows and summing corresponding columns in the next layer. (b) Extracting lightweight principal features by averaging the normalization layer’s affine transformation parameters.}
    \label{fig:insights}
\end{figure}

\noindent\textbf{Insight 1. Leveraging teacher model weights for efficient knowledge transfer.}
Well-trained model weights encapsulate substantial knowledge, and their effective utilization can significantly enhance knowledge transfer efficiency. Our analysis reveals that traditional distillation methods, which rely solely on teacher model guidance, underperform on mainstream person ReID datasets. In contrast, our approach employs cluster centers of the teacher model's weight rows combined with a progressive strategy to efficiently transfer knowledge to the weight chain.

\begin{figure*}[t]
    \centering
    \includegraphics[width=\linewidth]{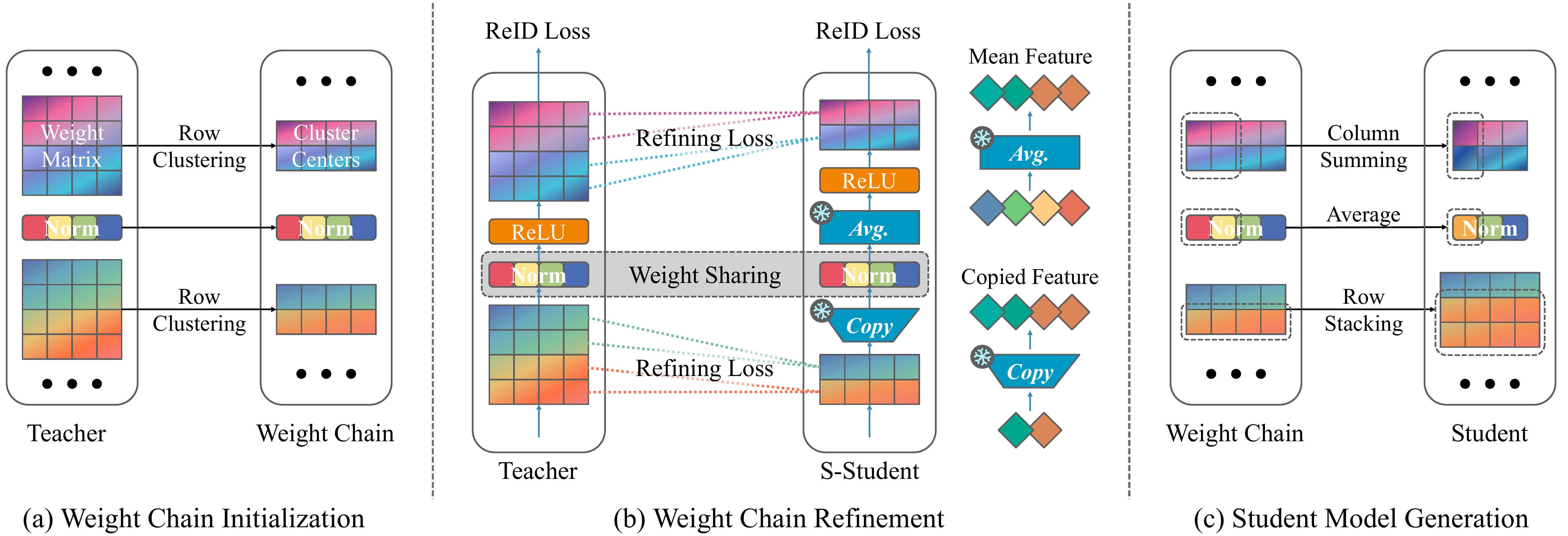}
    \caption{An overview framework of the proposed method.
    (a) Cluster the weight rows within each layer of the teacher model, using cluster centers as initial weight chain rows.
    (b) Train the teacher model and the S-Student constrained by the weight chain, using Refining loss and ReID loss to infuse core knowledge into the weight chain.
    (c) Reuse and stack the refined rows proportionally to achieve the required layer width for the student model. Average the $(\gamma,\beta)$ pairs within each normalization layer and sum the column weights of each subsequent layer accordingly to generate the student model. This is an O(1) operation that does not require additional computation.}
    \label{fig:method_process}
\end{figure*}

\noindent\textbf{Insight 2. Neural identity transformation.}  
In a neural network layer with multiple identical rows, if we ignore the subsequent normalization layers, merging these rows into one and summing their corresponding columns in the next layer preserves the network's function, as shown in Figure \hyperref[fig:insights]{3a}. This matrix multiplication property allows progressive refinement of knowledge into core weight rows, enabling the student model to approximate the teacher model's function.  
In practice, normalization layers must be addressed. Each feature dimension is associated with an affine transformation pair $(\gamma, \beta)$, and identical rows may correspond to distinct pairs, representing diverse feature transformations. To reduce feature dimensions by merging identical rows, their corresponding $(\gamma, \beta)$ pairs must also be merged. As shown in Figure \hyperref[fig:insights]{3b}, averaging these pairs retains critical information, leveraging the representativeness and minimal distance within clusters to enhance fine-tuning efficiency.

\noindent\textbf{Insight 3. Weight chain as a bridge between teacher and student models.}
The weight chain serves as a knowledge carrier, expandable to create student models ranging from the teacher model's size at maximum to its own width at minimum. Each layer's width in these student models can vary continuously within this range, positioning the weight chain as a bridge connecting the teacher model to a spectrum of student models.
At the bridge's ends lie the teacher model and the smallest expandable student model. By extending the weight chain to these two endpoint models and training with ReID loss, gradients propagate back to the weight chain. This approach trains only the two endpoint models while providing indirect training for all intermediate models. Like lifting a string of beads by their ends, training the endpoints elevates all intermediate models in between.


\subsection{Refinement of the Weight Chain}

\subsubsection{Row Clustering for the Weight Chain Initialization}
\label{sec:weight_chain_init}

The weight matrix of the teacher model encapsulates rich knowledge, which effectively initializes the weight chain, accelerating the refinement process. As illustrated in Figure \hyperref[fig:method_process]{4a}, at each layer, we cluster the weight rows of the teacher model, setting the number of clusters to match the width of the weight chain. The cluster centers, being representative and exhibiting smaller intra-cluster distances, initialize the corresponding rows in the weight chain, facilitating progressive refinement and enhancing knowledge transfer.


In layers with residual connections, where outputs from multiple layers are aggregated, we cluster these layers together to align the number of distinct feature dimensions with the number of weight chain rows in each layer. This enables us to initialize student models by summing the column weights in subsequent layers, effectively mimicking the functionality of the teacher model. 

For normalization layers, we directly share affine transformation weights between the weight chain and the teacher model, enabling a single row in the weight chain to correspond to multiple affine transformation pairs. While these parameters are few, they are critical for generating models of varying sizes through different levels of merging. This design also prevents student models from representing identical functions and stagnating in the same local optima.


\subsubsection{Progressive Refinement of the Weight Chain}

After obtaining teacher weight centers as initial weight chain rows, further optimization is needed because these centers, while representative, may not fully capture shared functionalities and can be affected by noisy weights.
We employ a progressive refinement process by simultaneously training the teacher model and the smallest student model (S-Student), both constrained by the weight chain. The S-Student shares the same width as the weight chain, enabling the effective transfer of shared knowledge from the teacher model to the weight chain throughout the training process.
Serving as a bridge, the weight chain connects the teacher model with student models of all widths. By training only the two models at either end of this bridge—the teacher model and the S-Student—we can achieve strong performance across all intermediate student models.

Specifically, we use Mean Squared Error (MSE) loss to tighten the clusters, aligning the weight rows more closely in the same cluster within the teacher model:
\begin{equation}\label{loss_refine}
\mathcal{L}_{ref} = \frac{1}{L}\sum_{\substack{1\leq l\leq L}}\frac{1}{M_l}\sum_{\substack{1 \leq j \leq M_l \\ k \in \mathcal{I}_l^{(j)}}}(\mathcal{F}_{l, k}-\mathcal{F}^{C}_{l, j})^2,
\end{equation}
where $\mathcal{F}_{l, k}$ and $\mathcal{F}^{C}_{l, j}$ are the weight rows in the $l$-th layer of the teacher model and the weight chain, respectively, and $\mathcal{I}_l^{(j)}$ indexes the clustered teacher weight rows associated with the $j$-th row of the weight chain.
To preserve essential ReID knowledge during refinement, we train the teacher model using both ID loss ($\mathcal{L}_{id}$) and hard triplet loss ($\mathcal{L}_{tri}$). 
The overall loss on the teacher model, $\mathcal{L}_{T}$, is given by:
\begin{equation}\label{loss_AncNet}
\mathcal{L}_{T} = \mathcal{L}_{id}(\boldsymbol{p}^T)+\mathcal{L}_{tri}(\boldsymbol{f}^{T}),
\end{equation}
where $\boldsymbol{p}^T$ and $\boldsymbol{f}^{T}$ are the logits and features generated by the teacher model, respectively. The overall loss function on the S-Student, $\mathcal{L}_{S}$, is given by:
\begin{equation}\label{loss_GeneNet}
\mathcal{L}_{S} = \mathcal{L}_{id}(\boldsymbol{p}^S)+\mathcal{L}_{tri}(\boldsymbol{f}^{S}),
\end{equation}
where $\boldsymbol{p}^S$ and $\boldsymbol{f}^{S}$ are the logits and features generated by the S-Student. It is important to note that the gradients derived from the losses applied to the S-Student are backpropagated to the weight chain, thereby optimizing it.

The total loss, $\mathcal{L}$, integrates the refining loss $\mathcal{L}_{ref}$ with ReID losses $\mathcal{L}_{T}$ and $\mathcal{L}_{S}$ to optimize the weight chain:
\begin{equation}\label{eq:total_loss}
\mathcal{L} =  \mathcal{L}_{T} + \mathcal{L}_{S}+\alpha\mathcal{L}_{ref}.
\end{equation}
In ViTs, $\alpha$ is a progressive coefficient set as $\alpha=\frac{iter}{n\_iter}$, with $iter$ being the current iteration and $n\_iter$ the total iterations. For CNNs, $\alpha$ is fixed at 1.


\subsection{Student Generation with the Weight Chain}


In this section, we illustrate the student model generation process using the $l$-th layer of the student model as an example, as illustrated in Figure \hyperref[fig:method_process]{4c}. In the teacher model, the $l$-th layer has $N_{l}$ rows, while the weight chain’s $l$-th layer has $M_l$ refined rows. If the student model requires $C_l$ rows in this layer ($M_l \leq C_l \leq N_l$), we stack refined rows in the $l$-th layer of the weight chain proportionally to the number of corresponding teacher weight rows to achieve $C_l$ rows. This results in a student-weight chain-teacher rows matcher, where each student row corresponds to multiple teacher rows. We then sum the weights of the corresponding columns in subsequent layers to preserve similar functionality as before merging.
For the followed normalization $(\gamma, \beta)$ pairs associated with the same student row, we average these pairs to initialize the corresponding normalization layer in the student model. This averaging retains the most essential information from the refined features, enabling enhanced optimization in downstream scenarios.

\section{Experiments}

\subsection{Experimental Designs}
\noindent\textbf{Datasets and evaluation metrics.}
We evaluate on three prominent person ReID benchmarks: Market1501 \cite{zheng_scalable_2015}, MSMT17 \cite{wei_person_2018}, and CUHK03 \cite{li_deepreid_2014}, which are referred to as M, MS, and C, respectively. Evaluation metrics include Cumulative Matching Characteristic (CMC) and mean Average Precision (mAP), which are standard for comprehensive performance assessment.

\noindent\textbf{Teacher and student models.}
We leverage pre-trained ResNet50 \cite{he2016deep} from LUPerson \cite{fu2021unsupervised} for CNN architecture and pre-trained ViT-B/ViT-S from PASS \cite{zhu2022pass} for ViT architecture, fine-tuning them on person ReID datasets as teacher models. For ResNet50, we design six progressively scaled student models (Res-50-S1 to S6), each doubling its predecessor's parameters, while establishing two student models each for ViT-B (ViT-B-S1, S2) and ViT-S (ViT-S-S1, S2), with detailed specifications provided in the appendix.

\noindent\textbf{Comparison methods.}
Our study compares training from scratch, knowledge distillation (using KD++ \cite{wang2023improving} for feature alignment and norm enhancement), model pruning (evaluating DepGraph \cite{fang_depgraph_2023} for dependency-based group sparsity and fine-tuning), weight selection \cite{xuInitializingModelsLarger2023} (referred to as WTSel), and our OSKT, systematically assessing student model performance across these methods.

\begin{table}[t]
\caption{Results of transferring knowledge from ResNet50 to student models on the Market1501 dataset.
}
\label{tab:res_sft}
\centering
\setlength{\tabcolsep}{5.7pt}
\begin{tabular}{lccccccc} 
\toprule[1.5pt] 
\multirow{2}{*}{Method} & \multicolumn{2}{c}{Res-50-S1} & \multicolumn{2}{c}{Res-50-S3} & \multicolumn{2}{c}{Res-50-S5} \\
\cmidrule[0.5pt](lr){2-3} \cmidrule[0.5pt](lr){4-5} \cmidrule[0.5pt](lr){6-7}
& mAP & R1 & mAP & R1 & mAP & R1 \\
\midrule[0.5pt] 

Scratch & 30.6 & 53.3 & 47.4 & 70.9 & 60.7 & 80.9 \\
WTSel & 48.5 & 72.3 & 63.0 & 81.5 & 84.1 & 93.0 \\
KD++ & 41.5 & 65.7 & 59.9 & 80.4 & 71.4 & 86.9 \\
DepGraph & 61.3 & 80.7 & 83.2 & 92.7 & 86.3 & 94.3 \\
\rowcolor{gray!40}\textbf{OSKT} & \textbf{75.7} & \textbf{89.4} & \textbf{84.7} & \textbf{93.3} & \textbf{87.6} & \textbf{94.5} \\

\bottomrule[1.5pt] 
\end{tabular}
\end{table}

\begin{table}[t]
\caption{
Results of transferring knowledge from ViT-B and ViT-S to their respective two student models on the Market-1501 dataset. Results are reported in terms of mAP/Rank-1.
}
\label{tab:vit_sft}
\centering
\setlength{\tabcolsep}{4.1pt}
\begin{tabular}{lcccc} 
\toprule[1.5pt] 
Method & ViT-S-S1 & ViT-S-S2 & ViT-B-S1 & ViT-B-S2 \\
\midrule[0.5pt] 

Scratch & 13.9/23.9 & 18.3/31.2 & 22.2/37.4 & 21.9/36.3 \\
WTSel & 41.0/60.5 & 54.8/75.1 & 16.8/31.7 & 58.7/78.2 \\
KD++ & 22.6/38.7 & 25.0/41.2 & 25.9/41.7 & 28.2/44.4 \\
DepGraph & 56.5/74.1 & 69.0/84.2 & 15.3/30.1 & 81.5/91.7 \\
\rowcolor{gray!40}\textbf{OSKT} & \textbf{74.2}/\textbf{87.1} & \textbf{77.2}/\textbf{89.0} & \textbf{81.6}/\textbf{91.9} & \textbf{82.9}/\textbf{92.4} \\

\bottomrule[1.5pt] 
\end{tabular}
\end{table}

\begin{table*}[t]
\caption{
Results of transferring knowledge from ResNet50 to student models. K.T.: knowledge transfer using the source dataset. F.T.: fine-tune on the downstream dataset. $k$: \# of student models. $m$: \# of weight chains refined from the teacher model. In this table, $m=3$.
}
\label{tab:res_da}
\centering
\setlength{\tabcolsep}{5.1pt}
\begin{tabular}{p{0.4cm}lcccccccccccccc} 
\toprule[1.5pt] 
& \multirow{2}{*}{Method} & \multicolumn{2}{c}{Epoch} & \multicolumn{2}{c}{Res-50-S1} & \multicolumn{2}{c}{Res-50-S2} & \multicolumn{2}{c}{Res-50-S3} & \multicolumn{2}{c}{Res-50-S4} & \multicolumn{2}{c}{Res-50-S5} & \multicolumn{2}{c}{Res-50-S6} \\
\cmidrule[0.5pt](lr){3-4} \cmidrule[0.5pt](lr){5-6} \cmidrule[0.5pt](lr){7-8} \cmidrule[0.5pt](lr){9-10} \cmidrule[0.5pt](lr){11-12} \cmidrule[0.5pt](lr){13-14} \cmidrule[0.5pt](lr){15-16}
& & K.T. & F.T. & mAP & R1 & mAP & R1 & mAP & R1 & mAP & R1 & mAP & R1 & mAP & R1  \\
\midrule[0.5pt] 

\multirow{4}*{\rotatebox{90}{MS$\rightarrow$M}} & WTSel & - & 100$\cdot k$ & 41.3 & 64.0 & 54.6 & 76.5 & 60.1 & 80.2 & 73.1 & 88.5 & 77.4 & 90.7 & \textbf{86.1} & 94.0 \\
& KD++ & 200$\cdot k$ & 100$\cdot k$ & 54.6 & 76.7 & 63.2 & 82.5 & 71.2 & 87.5 & 77.5 & 90.5 & 78.9 & 91.0 & 80.9 & 92.2 \\
& DepGraph & 100$\cdot k$ & 100$\cdot k$ & 61.6 & 81.0 & \textbf{75.0} & 89.2 & 77.3 & 90.3 & 82.4 & \textbf{92.8} & 82.3 & 92.7 & 82.1 & 92.5 \\
&\cellcolor{gray!40}\textbf{OSKT} &\cellcolor{gray!40}100$\cdot m$ &\cellcolor{gray!40}100$\cdot k$ &\cellcolor{gray!40}\textbf{72.3} &\cellcolor{gray!40}\textbf{87.5} &\cellcolor{gray!40}74.4 &\cellcolor{gray!40}\textbf{89.6} &\cellcolor{gray!40}\textbf{82.6} &\cellcolor{gray!40}\textbf{92.6} &\cellcolor{gray!40}\textbf{82.9} &\cellcolor{gray!40}92.7 &\cellcolor{gray!40}\textbf{85.6} &\cellcolor{gray!40}\textbf{93.9} &\cellcolor{gray!40}\textbf{86.1} &\cellcolor{gray!40}\textbf{94.1} \\

\midrule[0.5pt] 
\multirow{4}*{\rotatebox{90}{MS$\rightarrow$C}} & WTSel & - & 100$\cdot k$ & 16.6 & 15.5 & 24.8 & 24.6 & 22.2 & 21.3 & 38.9 & 40.1 & 45.1 & 45.5 & 68.9 & \textbf{71.8} \\
& KD++ & 200$\cdot k$ & 100$\cdot k$ & 26.8 & 27.0 & 34.3 & 35.4 & 41.6 & 42.9 & 52.0 & 53.3 & 53.6 & 55.1 & 56.1 & 57.6 \\
& DepGraph & 100$\cdot k$ & 100$\cdot k$ & 31.3 & 31.7 & 49.0 & 51.5 & 54.8 & 57.8 & 62.4 & 64.4 & 65.0 & 68.7 & 65.8 & 69.9 \\
&\cellcolor{gray!40}\textbf{OSKT} &\cellcolor{gray!40}100$\cdot m$ &\cellcolor{gray!40}100$\cdot k$ &\cellcolor{gray!40}\textbf{45.7} &\cellcolor{gray!40}\textbf{47.3} &\cellcolor{gray!40}\textbf{49.2} &\cellcolor{gray!40}\textbf{52.2} &\cellcolor{gray!40}\textbf{62.5} &\cellcolor{gray!40}\textbf{65.1} &\cellcolor{gray!40}\textbf{63.1} &\cellcolor{gray!40}\textbf{65.8} &\cellcolor{gray!40}\textbf{68.6} &\cellcolor{gray!40}\textbf{71.7} &\cellcolor{gray!40}\textbf{69.8} &\cellcolor{gray!40}71.6 \\

\midrule[0.5pt] 
\multirow{4}*{\rotatebox{90}{M$\rightarrow$C}} & WTSel & - & 100$\cdot k$ & 18.0 & 15.9 & 27.4 & 27.1 & 28.0 & 27.9 & 45.3 & 47.4 & 55.0 & 56.9 & 65.7 & 68.3 \\
& KD++ & 200$\cdot k$ & 100$\cdot k$ & 21.1 & 21.1 & 31.8 & 32.1 & 37.5 & 37.5 & 46.1 & 47.3 & 50.3 & 52.1 & 55.9 & 58.4 \\
& DepGraph & 100$\cdot k$ & 100$\cdot k$ & 23.4 & 22.7 & 47.7 & 49.9 & 53.4 & 56.4 & \textbf{65.1} & \textbf{68.2} & 65.8 & 68.3 & 67.0 & 70.5 \\
&\cellcolor{gray!40}\textbf{OSKT} &\cellcolor{gray!40}100$\cdot m$ &\cellcolor{gray!40}100$\cdot k$ &\cellcolor{gray!40}\textbf{44.6} &\cellcolor{gray!40}\textbf{47.4} &\cellcolor{gray!40}\textbf{47.9} &\cellcolor{gray!40}\textbf{50.9} &\cellcolor{gray!40}\textbf{60.9} &\cellcolor{gray!40}\textbf{64.1} &\cellcolor{gray!40}63.5 &\cellcolor{gray!40}66.5 &\cellcolor{gray!40}\textbf{68.0} &\cellcolor{gray!40}\textbf{70.6} &\cellcolor{gray!40}\textbf{69.5} &\cellcolor{gray!40}\textbf{72.3} \\

\bottomrule[1.5pt] 
\end{tabular}
\end{table*}

\begin{table}[t]
\caption{Results of transferring knowledge from ViT-S and ViT-B to their respective student models under three across-scenario settings. Results are reported in terms of mAP/Rank-1.}
\label{tab:vit_da}
\centering
\setlength{\tabcolsep}{2.95pt}
\begin{tabular}{ll|cc|cc}
\toprule[1.5pt] 
& Method & ViT-S-S1 & ViT-S-S2 & ViT-B-S1 & ViT-B-S2 \\
\midrule[0.5pt] 

\multirow{4}*{\rotatebox{90}{MS$\rightarrow$M}} & WTSel & 42.8/62.9 & 55.4/75.3 & 15.8/30.5 & 60.0/79.1 \\
& KD++ & 38.1/55.9 & 41.2/60.7 & 40.2/59.3 & 40.7/59.4 \\
& DepGraph & 65.5/82.1 & 74.6/86.9 & 63.7/80.7 & \textbf{83.3}/\textbf{92.1} \\
&\cellcolor{gray!40}\textbf{OSKT} &\cellcolor{gray!40}\textbf{76.5}/\cellcolor{gray!40}\textbf{88.7} &\cellcolor{gray!40}\textbf{79.0}/\cellcolor{gray!40}\textbf{89.7} &\cellcolor{gray!40}\textbf{82.1}/\cellcolor{gray!40}\textbf{91.9} &\cellcolor{gray!40}83.0/\cellcolor{gray!40}92.0 \\

\midrule[0.5pt] 
\multirow{4}*{\rotatebox{90}{MS$\rightarrow$C}} & WTSel & 1.6/0.9 & 20.9/19.9 & 3.0/2.5 & 24.2/23.1 \\
& KD++ & 15.8/14.1 & 17.9/16.6 & 18.4/17.1 & 20.5/18.7 \\
& DepGraph & 46.7/47.9 & 59.4/61.3 & 41.1/43.1 & \textbf{71.0}/\textbf{74.5} \\
&\cellcolor{gray!40}\textbf{OSKT} &\cellcolor{gray!40}\textbf{61.2}/\cellcolor{gray!40}\textbf{63.5} &\cellcolor{gray!40}\textbf{63.9}/\cellcolor{gray!40}\textbf{66.4} &\cellcolor{gray!40}\textbf{69.4}/\cellcolor{gray!40}\textbf{72.4} &\cellcolor{gray!40}70.3/\cellcolor{gray!40}73.0 \\

\midrule[0.5pt] 
\multirow{4}*{\rotatebox{90}{M$\rightarrow$C}} & WTSel & 1.7/1.4 & 20.8/19.5 & 2.5/1.9 & 5.2/4.5 \\
& KD++ & 10.4/9.0 & 11.6/10.4 & 12.2/10.1 & 13.9/12.1 \\
& DepGraph & 40.3/41.4 & \textbf{51.7}/52.7 & 6.2/5.2 & \textbf{63.1}/\textbf{65.6} \\
&\cellcolor{gray!40}\textbf{OSKT} &\cellcolor{gray!40}\textbf{49.3}/\cellcolor{gray!40}\textbf{50.9} &\cellcolor{gray!40}51.3/\cellcolor{gray!40}\textbf{52.9} &\cellcolor{gray!40}\textbf{59.4}/\cellcolor{gray!40}\textbf{61.7} &\cellcolor{gray!40}61.9/\cellcolor{gray!40}64.4 \\

\bottomrule[1.5pt] 
\end{tabular}
\end{table}

\begin{table}[t]
\caption{
Cost of obtaining student models of ViT-S and ViT-B. K.T.: \# of epochs for knowledge transfer using the source dataset. F.T.: \# of epochs for fine-tuning on the downstream dataset. $k$: \# of student models. $m$: \# of distinct-width weight chains refined from teacher models. In Table \ref{tab:vit_da}, $m=1$ for both ViT-S and ViT-B.
}
\label{tab:vit_da_cost}
\centering
\setlength{\tabcolsep}{9.5pt}
\begin{tabular}{l|cc|cc} 
\toprule[1.5pt] 
\multirow{2}{*}{Method}
& \multicolumn{2}{c|}{Teacher: ViT-S} 
& \multicolumn{2}{c}{Teacher: ViT-B} \\
\cmidrule[0.5pt](lr){2-3} \cmidrule[0.5pt](lr){4-5} 
& K.T. & F.T. & K.T. & F.T. \\
\midrule[0.5pt] 

WTSel & - & 120$\cdot k$ & - & 120$\cdot k$  \\
KD++ & 120$\cdot k$ & 120$\cdot k$ & 120$\cdot k$ & 120$\cdot k$ \\
DepGraph & 120$\cdot k$ & 120$\cdot k$ & 120$\cdot k$ & 120$\cdot k$ \\
\rowcolor{gray!40}\textbf{OSKT} & 120$\cdot m$ & 120$\cdot k$ & 120$\cdot m$ & 120$\cdot k$ \\

\bottomrule[1.5pt] 
\end{tabular}
\end{table}

\subsection{Knowledge Transfer in a Single Scenario}
We assess the efficacy of knowledge transfer from a teacher model to student models using diverse methods on a single dataset, subsequently comparing their test performance on the same dataset.
The results in Table \ref{tab:res_sft} and Table \ref{tab:vit_sft} demonstrate that our method effectively leverages the teacher model's weights for knowledge transfer.

\subsection{Knowledge Transfer across Scenarios}
Cross-scene knowledge transfer initially leverages a person re-identification dataset to perform knowledge transfer and obtain a student model, followed by fine-tuning and evaluating the student model on a distinct downstream scene. This is a common practice in real-world applications, as it is often difficult to quickly obtain large amounts of labeled data for training in downstream scenarios.
Since knowledge transfer typically involves transferring from a larger dataset to a smaller one, we designed three cross-scenario transfer settings: from MSMT17 to Market1501 and CUHK03, as well as from Market1501 to CUHK03.
For downstream training, we employ the BoT \cite{luo2019bag} framework for ResNet and adhere to the PASS \cite{zhu2022pass} setup for ViT. As shown in Tables \ref{tab:res_da} and \ref{tab:vit_da}, our method consistently surpasses others, especially as the student models' sizes decrease.

\subsubsection{Convergence Speed}

\begin{figure}[t] 
    \centering
    \includegraphics[width=\linewidth]{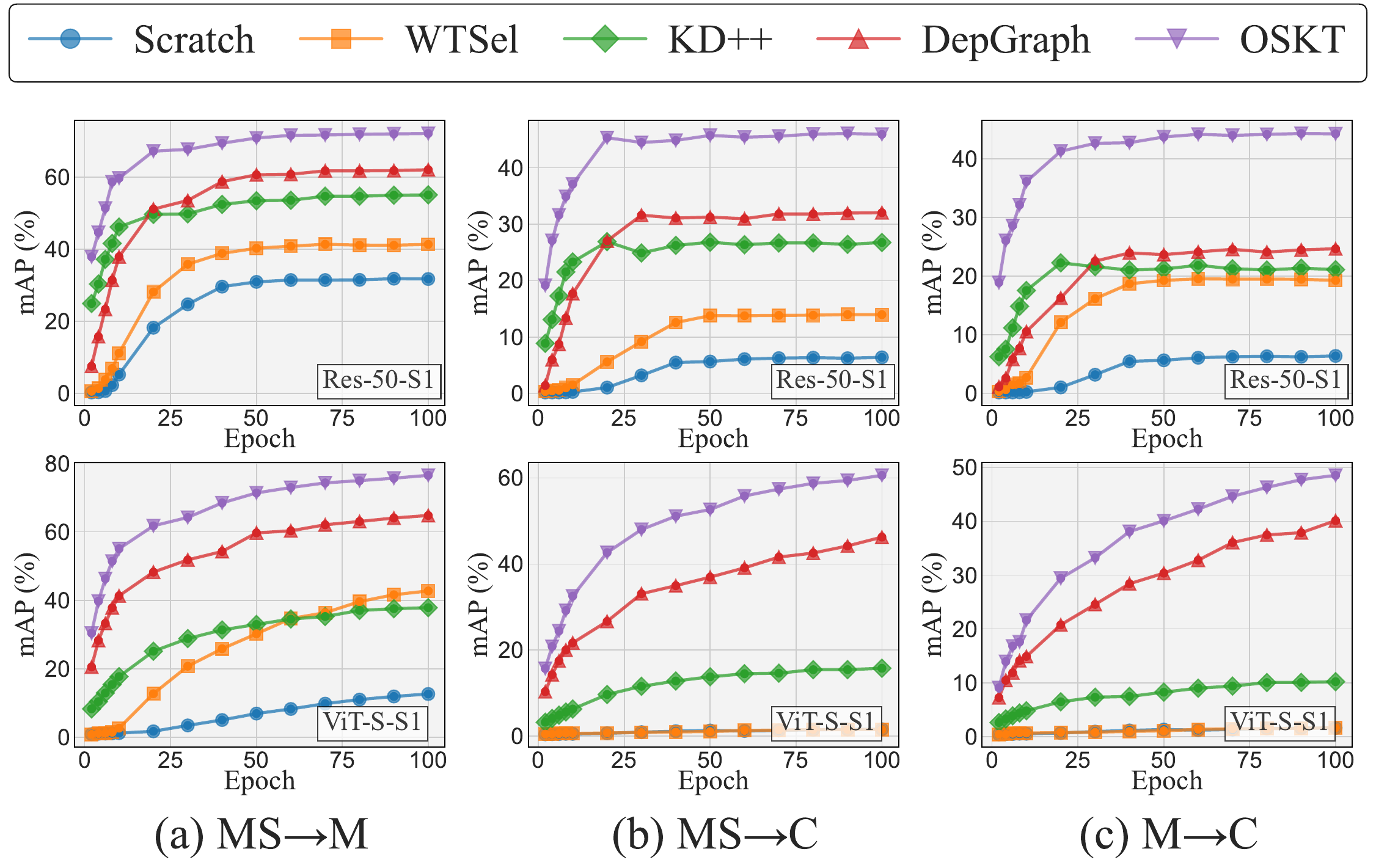}
    \caption{Convergence curves of student models generated by our OSKT, compared to other approaches across diverse scenarios.}
    \label{fig:speed_comparison}
\end{figure}

In Figure \ref{fig:speed_comparison}, we present the validation mAP curves for student models over the first 100 epochs on downstream datasets. Our method consistently surpasses others, offering reduced computational costs and enhanced efficiency.

\subsubsection{Few-shot Setting}

\begin{figure}[t]
    \centering
    \includegraphics[width=\linewidth]{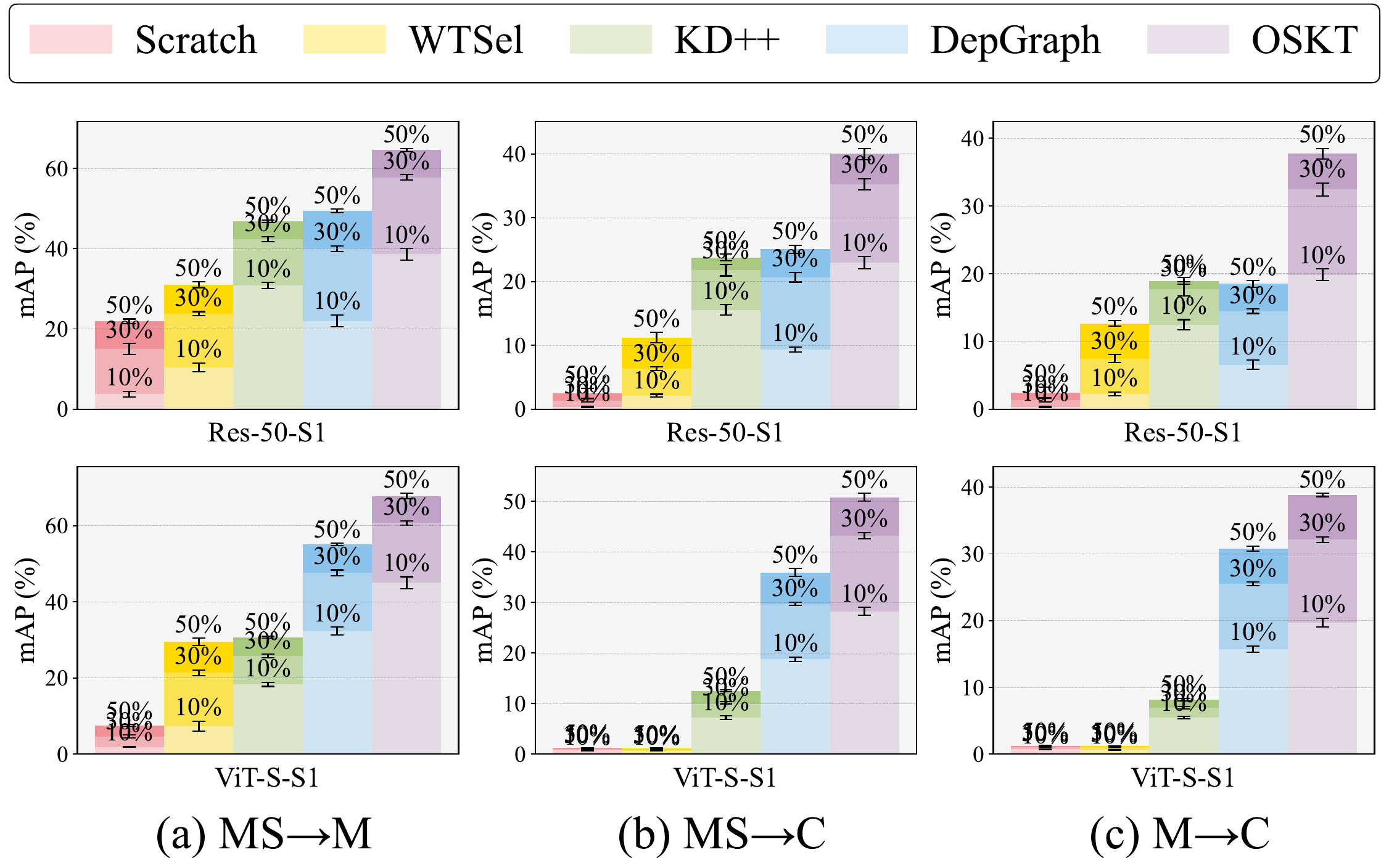}
    \caption{Performance comparison under few-shot settings, with each sampling ratio evaluated using five random seeds.}
    \label{fig:fewshot_setting}
\end{figure}

In Figure \ref{fig:fewshot_setting}, we present the results under settings using 10\%, 30\%, and 50\% of IDs on downstream datasets.
To ensure the robustness of the experimental results, we perform five random samplings of IDs for each few-shot setting.

\subsubsection{Combined with Lightweight ReID Architectures}
Our method demonstrates strong compatibility with modern lightweight ReID models. As shown in Table \ref{tab:light_weight}, we validate its effectiveness using two state-of-the-art architectures: OSNet \cite{zhouOmniScaleFeatureLearning2019} and MSINet \cite{guMSINetTwinsContrastive2023}. We adopt the full-size models as teachers and construct two smaller variants for each model with width multipliers of $0.25$ and $0.5$. Pre-training is performed on MSMT17, with evaluation on Market1501 and CUHK03, comparing results from scratch training, pre-training, and our proposed method.
Notably, both OSNet and MSINet employ a lightweight building block comprising a point-wise $1\times1$ convolution followed by a depth-wise $3\times3$ convolution, which we cluster as a single unit for weight chain initialization and refinement.

\begin{table}[t]
\caption{
Results of modern ReID models initialized with various methods. Pre-training dataset: MSMT17. $\beta$: width multiplier.
}
\label{tab:light_weight}
\centering
\setlength{\tabcolsep}{4.4pt}
\begin{tabular}{lccccccc} 
\toprule[1.5pt] 

& \multirow{2}{*}{$\beta$} & \multirow{2}{*}{Params} & \multirow{2}{*}{Method} & \multicolumn{2}{c}{Market1501} & \multicolumn{2}{c}{CUHK03} \\
\cmidrule[0.5pt](lr){5-6} \cmidrule[0.5pt](lr){7-8}
& & & & mAP & R1 & mAP & R1 \\
\midrule[0.5pt] 

\multirow{6}*{\rotatebox{90}{OSNet \cite{zhouOmniScaleFeatureLearning2019}}} & \multirow{3}{*}{0.25} & \multirow{3}{*}{0.15M} & Scratch & 78.3 & 90.7 & 48.4 & 51.5 \\
& & & Pretrain & 80.0 & 91.6 & 58.0 & 61.8 \\
& & & \cellcolor{gray!40}\textbf{OSKT} & \cellcolor{gray!40}\textbf{81.8} & \cellcolor{gray!40}\textbf{92.5} & \cellcolor{gray!40}\textbf{60.6} & \cellcolor{gray!40}\textbf{63.6} \\
\cmidrule(l){2-8} 

& \multirow{3}{*}{0.5} & \multirow{3}{*}{0.56M} & Scratch & 83.7&93.4 & 57.9&60.6 \\
& & & Pretrain & 84.7 & 93.4 & 65.9 & 69.1 \\
& & & \cellcolor{gray!40}\textbf{OSKT} & \cellcolor{gray!40}\textbf{85.4} & \cellcolor{gray!40}\textbf{94.2} & \cellcolor{gray!40}\textbf{66.9} & \cellcolor{gray!40}\textbf{69.6} \\
\midrule[0.5pt] 

\multirow{6}*{\rotatebox{90}{MSINet \cite{guMSINetTwinsContrastive2023}}} & \multirow{3}{*}{0.25} & \multirow{3}{*}{0.17M} & Scratch & 81.2&92.7 & 53.3&55.7 \\
& & & Pretrain & 81.8&92.5 & 57.7&59.9 \\
& & & \cellcolor{gray!40}\textbf{OSKT} & \cellcolor{gray!40}\textbf{82.4} &\cellcolor{gray!40}\textbf{92.8} & \cellcolor{gray!40}\textbf{60.2} &\cellcolor{gray!40}\textbf{62.4} \\
\cmidrule(l){2-8} 

& \multirow{3}{*}{0.5} & \multirow{3}{*}{0.63M} & Scratch & 86.0&94.4 & 64.1&66.6 \\
& & & Pretrain & 87.4&94.6 & 70.9&73.4 \\
& & & \cellcolor{gray!40}\textbf{OSKT} & \cellcolor{gray!40}\textbf{87.9}&\cellcolor{gray!40}\textbf{95.0} & \cellcolor{gray!40}\textbf{72.8}&\cellcolor{gray!40}\textbf{75.6} \\

\bottomrule[1.5pt] 
\end{tabular}
\end{table}

\subsection{Ablation Study}

\subsubsection{Ablation Studies of Alternative Settings}

\begin{table*}[t]
\caption{Ablation of alternative settings in OSKT.
Results are reported in terms of mAP/Rank-1.
}
\label{tab:ablation}
\centering
\setlength{\tabcolsep}{5.4pt}
\begin{tabular}{l|cccccccc} 
\toprule[1.5pt] 
\multirow{2}{*}{Setting} & 
\multicolumn{4}{c}{Single Scenario: M} & 
\multicolumn{4}{c}{Across Scenarios: MS$\rightarrow$C} \\
\cmidrule[0.5pt](lr){2-5}\cmidrule[0.5pt](lr){6-9} & Res-50-S1 & Res-50-S2 & ViT-S-S1 & ViT-S-S2 & Res-50-S1 & Res-50-S2 & ViT-S-S1 & ViT-S-S2 \\
\midrule[0.5pt] 

Scratch & 30.6/53.3 & 41.5/65.1 & 13.9/23.9 & 18.3/31.2 & 6.1/4.8 & 9.1/6.7 & 1.5/0.5 & 2.3/1.4 \\
\midrule[0.5pt] 
(a) rand teacher & 54.7/76.5 & 58.2/78.9 & 33.0/50.5 & 33.9/52.1 & 28.2/28.2 & 30.8/30.9 & 14.5/13.0 & 15.8/14.1 \\
(b) rand cluster & 55.1/76.2 & 59.9/80.2 & 66.1/83.4 & 70.6/85.8 & 31.4/33.8 & 34.9/35.4 & 51.4/51.9 & 54.4/55.8 \\
(c) inverse cluster & 46.1/70.2 & 54.8/77.0 & 66.8/83.7 & 69.3/85.2 & 13.3/11.9 & 17.6/16.1 & 47.7/49.5 & 49.6/51.0 \\
(d) distance metric & 74.2/88.9 & 76.7/89.6 & 64.9/82.1 & 69.9/85.1 & 37.6/38.9 & 41.4/44.0 & 52.3/53.7 & 56.6/58.1 \\
(e) w/o refinement & 52.2/73.3 & 59.8/79.3 & 56.6/75.8 & 58.9/78.1 & 21.6/20.9 & 29.4/29.0  & 25.2/24.2 & 26.6/26.9 \\
(f) w/o $\mathcal{L}_{T}$ & 62.7/81.2 & 64.0/81.5 & 64.4/82.0 & 65.4/82.9 & 34.2/35.1 & 35.4/37.3 & 44.8/46.0 & 45.9/48.7 \\
(g) weight of $\mathcal{L}_{ref}$ & 75.6/\textbf{89.4} & 77.1/89.3 & 73.8/\textbf{87.3} & 75.2/87.9 & 45.5/\textbf{47.4} & 48.0/51.1 & 59.1/61.2 & 60.0/61.9 \\
\midrule[0.5pt] 
\rowcolor{gray!40}\textbf{OSKT} & \textbf{75.7}/\textbf{89.4} & \textbf{77.6}/\textbf{90.6} & \textbf{74.2}/87.1 & \textbf{77.2}/\textbf{89.0} & \textbf{45.7}/47.3 & \textbf{49.2}/\textbf{52.2} & \textbf{61.2}/\textbf{63.5} & \textbf{63.9}/\textbf{66.4} \\

\bottomrule[1.5pt] 
\end{tabular}
\end{table*}

We design ablation studies focusing on the initialization and refinement of the weight chain to provide further insights into our framework. The results are shown in Table \ref{tab:ablation}.

\noindent\textbf{Leveraging weights from the teacher model.}
We replace the method of utilizing weights from the teacher model with randomly initializing the parameters of the teacher model and then generating the student model according to Figure \ref{fig:method_process}. The results are shown as in (a) of Table \ref{tab:ablation}.

\noindent\textbf{Method for initializing the weight chain.}
We substitute the method of employing clustering to obtain weight centers for initializing the weight chain with two alternative approaches: random clustering and inverse clustering, as presented in (b) and (c) of Table \ref{tab:ablation}, respectively.

\noindent\textbf{Distance metric for clustering weight \textit{row}s.}
In the above experiments, for ResNet50, we clustered the weight rows of the teacher model based on Euclidean distance, while for ViT, we clustered based on cosine distance. By swapping these settings, we obtain the results shown in (d) of Table \ref{tab:ablation}.



\noindent\textbf{Progressive refinement of the weight chain.}
After initializing the weight chain using the teacher model, we omit the refinement step illustrated in Figure \hyperref[fig:method_process]{4b}, with the results shown in (e) of Table \ref{tab:ablation}.
During the refinement phase, we eliminate the loss term $\mathcal{L}_{T}$ from Equation \ref{eq:total_loss}. Consequently, the weight chain is no longer influenced by the loss of the teacher model but is solely driven by the gradients of the S-Student's loss $\mathcal{L}_S$. The corresponding results are presented in (f) of Table \ref{tab:ablation}.
In the above experiments, for ResNet50, the weight $\alpha$ of $\mathcal{L}_{ref}$ is set to $1$, while for ViTs, $\alpha=\frac{iter}{n\_iter}$ serves as a progressive weight. By swapping these settings, we obtain the results shown in (g) of Table \ref{tab:ablation}.

\subsubsection{Scalability Analysis of Weight Chains}

\begin{figure}[t]
    \centering
    \includegraphics[width=\linewidth]{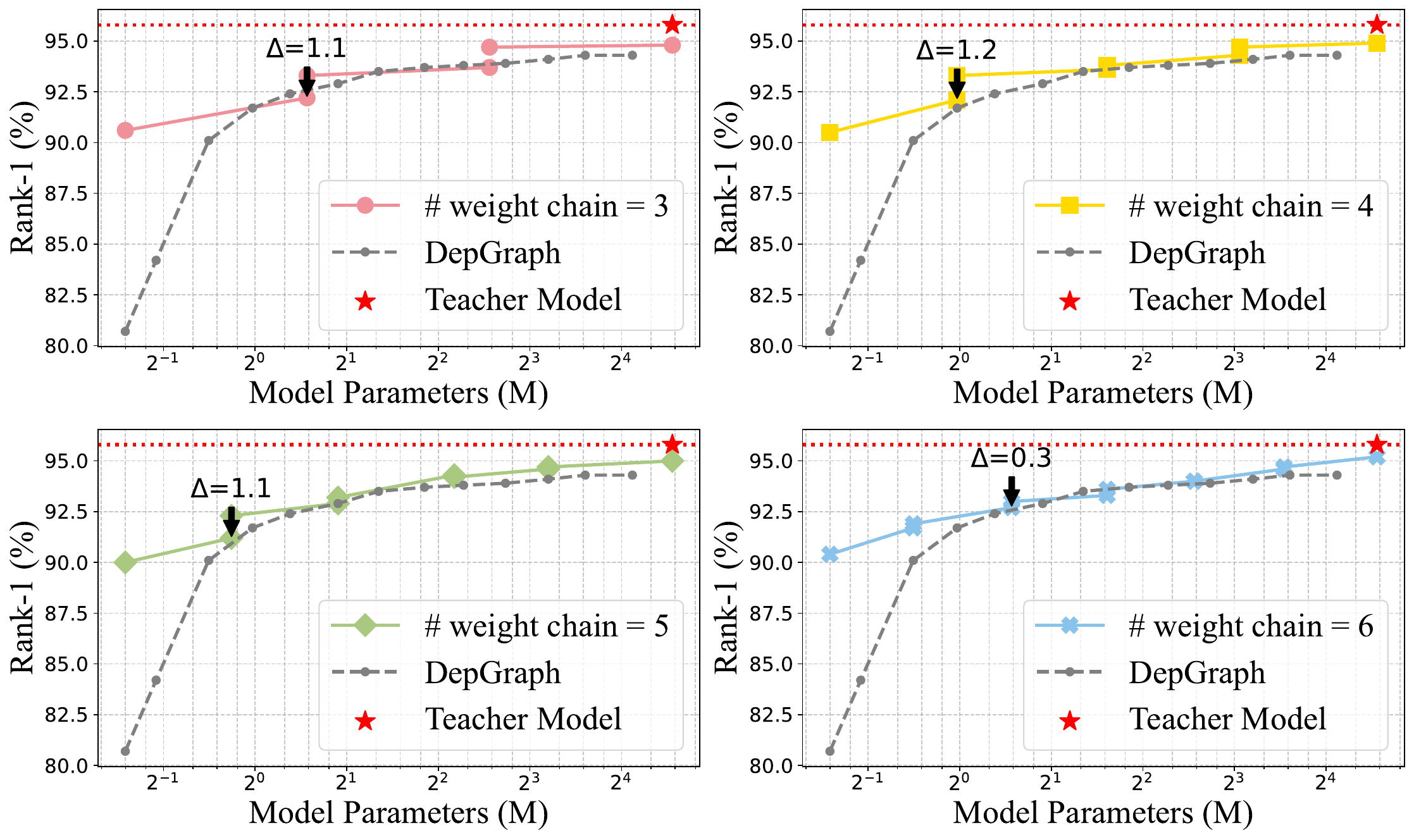}
    \caption{Scalability of weight chains refined from ResNet50 on Market1501. Each weight chain produces two models corresponding to its own width and the subsequent weight chain's width, with the maximum inter-stage discrepancy denoted by ``$\Delta$''.}
    \label{fig:weight_chain_scalability}
\end{figure}

When the size range of required student models is large, the weight chain's parameter count becomes significantly smaller than the teacher model's, leading to inefficiencies in generating larger student models. To address this, we analyze the scalability of the weight chain, determining how many chains are needed to maintain performance while covering a size range from small to teacher model size. Using ResNet50 with parameter counts spanning 1/64 to the teacher model's size, we present experimental results in Figure \ref{fig:weight_chain_scalability}.
We explore results using multiple weight chains, each responsible for a specific width range. For a size span $[a, b]$ and $s$ weight chains, the widths of the weight chains are generated as a geometric sequence, with the common ratio $x$ determined by solving $a \cdot x^{s} = b$. For instance, given a size span $inplanes \in [8,64]$ and $s = 3$, solving for $x$ yields $x = 2$, resulting in widths $8$, $16$, and $32$ for weight chains.

\subsubsection{Functional Analysis of Weight Chain Rows}
\begin{figure}[t]
    \centering
    \begin{subfigure}{0.49\linewidth}
        \centering
        \includegraphics[width=\linewidth]{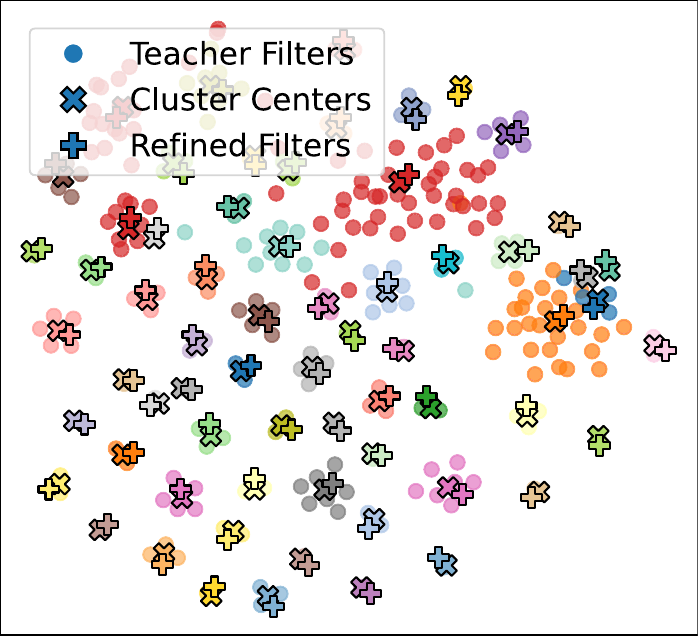}
        \caption{Cluster Visualization via T-SNE}
        \label{fig:cluster_vis}
    \end{subfigure}
    \hfill
    \begin{subfigure}{0.49\linewidth}
        \centering
        \includegraphics[width=\linewidth]{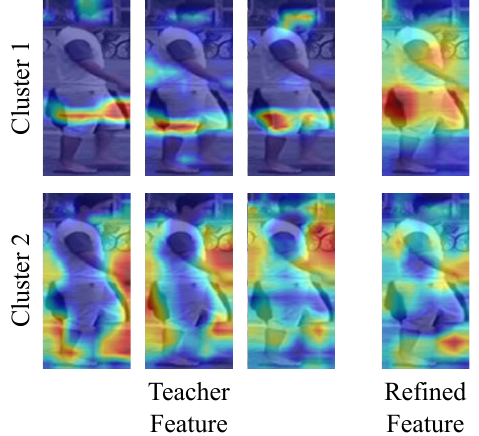}
        \caption{Original and Refined Feature}
        \label{fig:ori_ref_feature}
    \end{subfigure}
    \caption{(a) T-SNE visualization of the clustering and refining results for the layer3.5.conv2 in ResNet50. (b) Features extracted by filters from different clusters in the teacher model and their corresponding refined filters in the weight chain.}
    \label{fig:filter_clusters}
\end{figure}

We visualize the clustering and refining results of the weight rows, as illustrated in Figure \ref{fig:cluster_vis}. The features extracted from the weight rows of the teacher model and their corresponding refined weight rows in the weight chain are presented in Figure \ref{fig:ori_ref_feature}. Weight rows within the same cluster and their refined counterparts tend to activate similar features, as demonstrated by Cluster 1. Additionally, the refined weight rows can effectively transform noisy activation features into features containing discriminative pedestrian information, as exemplified by Cluster 2.

\section{Conclusion}

This study introduces a novel approach for efficiently generating high-performance person ReID models tailored to diverse resource constraints. Our OSKT meticulously refines core knowledge from the teacher model into a weight chain through a single computational pass, enabling the acquisition of resource-adaptive models without incurring additional computational costs in subsequent steps.


\section*{Acknowledgment}
This research was supported by the Jiangsu Science Foundation (BK20243012, BG2024036), the National Science Foundation of China (62125602, 62206052, U24A20324, 92464301), CAAI-Lenovo Blue Sky Research Fund, and the Fundamental Research Funds for the Central Universities (2242025K30024).

{
    \small
    \bibliographystyle{ieeenat_fullname}
    \bibliography{main}
}

\end{document}